\documentclass{article}

\usepackage{arxiv}

\usepackage[utf8]{inputenc}
\usepackage[T1]{fontenc}
\usepackage{microtype}
\usepackage{textcomp}

\usepackage{amsmath}
\usepackage{amsfonts}
\usepackage{amssymb}
\usepackage{bm}
\usepackage{algorithm}
\usepackage{algorithmic}

\usepackage{graphicx}
\graphicspath{ {./images/} }
\usepackage[caption=false,font=normalsize,labelfont=sf,textfont=sf]{subfig}
\usepackage{array}
\usepackage{booktabs}
\usepackage{multirow}
\usepackage{stfloats}
\usepackage{nicefrac}

\usepackage{verbatim}
\usepackage{lipsum}
\usepackage{orcidlink}

\usepackage{url}
\usepackage{hyperref}

\title{EcoFair: Energy-Efficient Inference Routing for Edge AI under Data Degradation\thanks{Manuscript created September 2026. This work was supported by the UK National Edge AI Hub under Grant Agreement EP/Y028813/1.}}

\author{
  \textbf{Mostafa Anoosha\,\orcidlink{0009-0000-2601-3692}\quad
  Dhavalkumar Thakker\,\orcidlink{0000-0003-4479-3500}\quad
  Kuniko Paxton\,\orcidlink{0009-0000-8897-9775}\quad
  Koorosh Aslansefat\,\orcidlink{0000-0001-9318-8177}} \\
  \textbf{Bhupesh Kumar Mishra\,\orcidlink{0000-0003-3430-8989}\quad
  Baseer Ahmad\,\orcidlink{0000-0001-9440-8514}\quad
  Rameez Raja Kureshi\,\orcidlink{0000-0002-2021-8053}} \\
  \vspace{0.2cm}
  \normalfont School of Digital and Physical Science, University of Hull \\
  Kingston upon Hull, HU6 7RX, U.K. \\
  \texttt{\{M.Anoosha, D.Thakker, k.Azuma-2021, K.Aslansefat,} \\
  \texttt{Bhupesh.Mishra, Baseer.Ahmad, R.Kureshi\}@hull.ac.uk}
}

\begin{document}
\maketitle

\begin{abstract}
    Medical edge-AI systems must operate under a difficult tension: delivering reliable diagnostic inference while running on devices with limited battery capacity, memory, and compute. In dermatology, this problem is amplified by real-world image degradation caused by smartphone capture, poor lighting, blur, compression, and heterogeneous edge sensors. To handle these degraded inputs, deploying a heavyweight model can improve reliability, but it rapidly increases the energy burden on resource-constrained devices. Conversely, always using a lightweight model saves energy but may be less reliable on ambiguous or degraded inputs. This paper introduces EcoFair, a vertically partitioned inference framework for dermatology classification in which image and tabular inputs remain local to edge clients while only learned modality-specific representations are transmitted for server-side fusion. EcoFair first processes each sample using a lightweight image encoder and then decides whether additional heavyweight computation is necessary. Escalation is triggered when the lightweight prediction exhibits high uncertainty, a narrow separation between safe and high-risk classes, or elevated metadata-derived risk from patient age and lesion location. Across HAM10000, BCN20000, and PAD-UFES-20, EcoFair is evaluated using multiple lightweight--heavy backbone pairings to quantify the trade-off between energy consumption, diagnostic performance, and worst-group malignant-case recall. Results show that EcoFair can reduce per-sample image-inference energy by up to 68\% relative to always using the heavyweight encoder, while selectively allocating additional computation under difficult data regimes to support inference reliability. Group-level analysis further shows configuration-dependent effects, with improvements in selected model--dataset settings and mixed behaviour in others.
\end{abstract}

\keywords{Edge AI \and Trustworthy AI \and Degradation-aware inference \and Distributed intelligence \and Data degradation \and Inference routing}

\section{Introduction}
\label{sec:introducion}
    
    The transition towards decentralised digital infrastructures has reshaped how sensitive information is processed and exchanged across ad hoc and edge networks. In regulated domains such as healthcare, where data use is constrained by frameworks including HIPAA (Health Insurance Portability and Accountability Act) and GDPR (General Data Protection Regulation), raw clinical data cannot be assumed to move freely into centralised repositories \cite{fl_privacy_healthcare_2025}. This is particularly important for image-based medical applications, where high-resolution images captured by mobile or IoT devices may contain identifiable biometric information while remaining essential for diagnosis. In the setting considered in this study, a dermatology image client processes skin-lesion images, a tabular client stores patient metadata, and a fusion server receives intermediate representations for multimodal inference. By exchanging learned embeddings rather than raw pixels or metadata, such architectures reduce raw-data sharing while preserving the possibility of distributed inference. 
    
    Vertically partitioned learning and inference architectures provide a practical mechanism for collaboration when distributed nodes hold complementary information about the same patient population \cite{liu2024vertical}. In such settings, each node can retain its raw inputs while contributing intermediate representations for downstream inference \cite{vepakomma2018split}. This design is particularly relevant when image and metadata features are distributed across separate holders \cite{chen2024metadata}. However, edge deployment introduces additional constraints beyond data locality. Mobile dermatoscopes, smartphone cameras, and clinical IoT devices may operate under variable lighting, limited battery capacity, intermittent connectivity, and heterogeneous compute resources \cite{edge_iot_healthcare_survey}. Existing techniques such as quantisation, pruning, and knowledge distillation can reduce computation, but may also weaken representational quality and downstream inference reliability \cite{liu2025survey}. In safety-critical medical benchmarks, efficiency therefore cannot be evaluated only through average accuracy or energy saving; it is also important to examine whether adaptive inference changes malignant-case behaviour across clinically relevant subgroups.
    
    EcoFair is designed to address these deployment challenges through a degradation-aware inference-routing framework. Rather than presenting an end-to-end vertically partitioned training pipeline, this work studies a simulated vertically partitioned inference architecture designed to operate under edge constraints. Raw image and tabular data remain local, only modality-specific embeddings are transmitted to the server, and a dynamic routing mechanism determines whether lightweight inference is sufficient or whether a heavier image encoder should be activated before multimodal fusion. To examine this behaviour beyond a single fixed backbone choice, EcoFair is evaluated using multiple pretrained lightweight and heavyweight image encoders, including MobileNetV2, MobileNetV3Small, ResNet50, DenseNet201, and EfficientNetB6. The overarching aim is to study resource-aware routing under degraded or ambiguous inputs while quantifying group-level malignant-case behaviour across datasets and backbone configurations.
    
    \textbf{Scope of this study.} EcoFair targets sensor-degradation resilience, defined narrowly as degraded-input detection and adaptive heavyweight escalation when lightweight edge inference may be unreliable. This scope is distinct from broader cyber-physical resilience. The evaluation is conducted in a simulated vertically partitioned inference setting with measured encoder-level energy costs. EcoFair reduces raw-data sharing through vertical partitioning, while formal guarantees such as differential privacy, encryption, secure aggregation, Byzantine robustness, adversarial attack tolerance, and end-to-end resilience to network churn remain outside the present scope. These dimensions are complementary to the proposed routing mechanism and remain future deployment-validation targets.
    
    Accordingly, this study is guided by the following research questions:
    
    \begin{enumerate}
        \item \textbf{RQ1:}\label{rq1} How can decentralised medical inference maintain resource-aware inference reliability at the edge without transmitting raw biometric data to a central server?
        \item \textbf{RQ2:}\label{rq2} How can dynamic routing incorporate metadata-derived risk to support escalation when lightweight predictions are uncertain or ambiguous under data degradation?
        \item \textbf{RQ3:}\label{rq3} To what extent does post-hoc routing affect subgroup-level malignant-case performance across datasets and backbone configurations without modifying the global training objective?
    \end{enumerate}
    
    In response to these questions, this paper makes the following contributions:

    \begin{itemize}
        \item To address RQ\hyperref[rq1]{1}, we propose a vertically partitioned inference framework that reduces raw-data sharing in which raw image and tabular data remain local and only modality-specific embeddings are exchanged. Within this setting, EcoFair introduces a resource-aware routing strategy that supports degradation-aware inference reliability while avoiding unnecessary heavyweight escalation and transmission overheads.
    
        \item To address RQ\hyperref[rq2]{2}, we combine predictive uncertainty with a rule-based metadata risk score based on patient age and lesion localisation, providing an interpretable routing signal that supports escalation for uncertain or higher-risk cases under degraded-input conditions.
    
        \item To address RQ\hyperref[rq3]{3}, we evaluate whether routing-based escalation changes group-level malignant-case behaviour post hoc, reporting both improvements and failure cases without treating routing as a fairness guarantee.
    \end{itemize}
    
    Unlike prior Edge AI approaches that primarily optimise computation or communication efficiency, EcoFair treats inference routing as a deployment-oriented mechanism for balancing resource use and degraded-input reliability. Its novelty lies in combining lightweight-first adaptive escalation, metadata-derived risk scoring, encoder-level energy analysis, and group-level malignant-case evaluation within a vertically partitioned dermatology inference setting. This allows routing to be evaluated not merely as an efficiency tool, but as a practical mechanism for responding to uncertain or degraded inputs.
    
    The rest of this paper is organised as follows. Section~\ref{sec:related_work} reviews related work. Section~\ref{sec:methodology} presents the EcoFair methodology, including the vertically partitioned inference setting, routing mechanism under edge constraints, and rule-based metadata risk design. Section~\ref{sec:experimental_evaluation} reports the experimental evaluation across resource use, classification performance, and group-level evaluation perspectives. Section~\ref{sec:discussion} discusses the main findings of this work. Section~\ref{sec:future_work} outlines limitations of the current study and directions for future research, and Section~\ref{sec:conclusion} concludes the paper.

\section{Related Work}
\label{sec:related_work}

    The deployment of robust, efficient, and trustworthy artificial intelligence (AI) at the edge has motivated substantial research across several intersecting areas of computer science and network engineering. To position EcoFair within this landscape, this section reviews relevant work on distributed data exchange, resource-aware system optimisation, and trustworthy medical AI.
    
    A central requirement in collaborative edge computing and ad hoc networks is balancing diagnostic reliability with strict on-device resource constraints. Recent works emphasise the necessity of adaptive offloading and intelligent resource management to guarantee high-reliability, low-latency services in constrained Internet of Things (IoT) environments \cite{sweidan2025rl}. Hierarchical and split-inference frameworks have also been proposed to dynamically partition neural workloads across edge and server components \cite{zhang2024coarse}. Vertically partitioned learning provides a related mechanism for collaboration when different nodes or institutions hold complementary attributes of the same individuals \cite{liu2024vertical}. This enables joint inference without direct raw-data pooling \cite{chen2024metadata}. It is also closely related to split-learning designs in which intermediate representations, rather than raw inputs, are exchanged across participating parties \cite{vepakomma2018split}. In this context, EcoFair is aligned with vertically partitioned inference settings that reduce direct raw-data exchange and are increasingly advocated for clinical use \cite{rieke2020future}, while focusing specifically on post-deployment optimisation under edge constraints rather than end-to-end federated training.
    
    The deployment of deep models in decentralised and edge environments introduces strict constraints related to energy, latency, and computational capacity \cite{chen2019deep}. Recent split-learning work has sought to mitigate these bottlenecks by dynamically regulating local computation under resource heterogeneity \cite{zhao2026split}. Prior studies have also explored model compression, lightweight architectures, pruning, quantisation, and knowledge distillation for resource-constrained inference \cite{han2022dynamic}. Related work has considered adaptive mechanisms for skewed or heterogeneous edge settings \cite{paxton2025skewness}. Other studies have examined convergence and system efficiency in distributed learning environments \cite{wang2020convergence}. Likewise, dynamic routing has emerged as a useful mechanism for selectively allocating computational effort based on input difficulty. This mitigates the computational waste of network ``overthinking'' on simpler samples, reserving heavier local compute capacity for degraded or highly uncertain inputs \cite{kaya2019shallow}. In a related direction, Qasim \textit{et al.} \cite{qasim2024transformer} examined the sustainability of federated networks and highlighted the trade-off between model complexity and energy cost. However, many routing strategies rely primarily on confidence thresholds that may be poorly calibrated in deep neural networks \cite{guo2017calibration}. EcoFair builds on this line of work by combining uncertainty-based routing with a rule-based metadata risk signal, while evaluating the resulting energy--performance trade-off under dermatology benchmark variation.
    
    In medical AI, trustworthiness requires more than average predictive performance. Deployed systems may show uneven behaviour across demographic groups, acquisition conditions, and lesion presentations \cite{seyyed2021underdiagnosis}. Broader fairness surveys have also highlighted how predictive systems can amplify existing biases if subgroup behaviour is not evaluated explicitly \cite{mehrabi2021survey}. In high-stakes settings, trustworthiness therefore requires attention to reliability, transparency, and subgroup-level assessment \cite{hasani2022trustworthy}. In dermatology, performance disparities may appear as reduced malignant-case sensitivity for minority subgroups or atypical lesion presentations \cite{daneshjou2022disparities}. Recent work has further emphasised the importance of measuring dermatology-model performance across diverse skin tones \cite{paxton2024measuring}. Clinician-facing studies also show that model outputs require careful evaluation before being treated as dependable decision support \cite{finlayson2021clinician}. At the same time, neurosymbolic methods have gained attention as a way to incorporate structured domain knowledge into learning systems and improve interpretability \cite{garcez2023neurosymbolic}. For example, the ZTID-IoV framework applied neurosymbolic design principles to interpretable zero-trust intrusion detection in decentralised vehicular networks \cite{ullah2025ztid}. EcoFair uses this motivation in a lightweight form by integrating a rule-based metadata risk score into the routing process, without framing the component as a full neurosymbolic reasoning system.
    
    Taken together, these limitations define the problem addressed in this paper: how to support distributed medical inference under edge deployment constraints while avoiding hidden degradation in malignant-case performance across relevant subgroups. Existing efficiency-oriented approaches in edge AI largely prioritise energy reduction, compression, or routing economy, but they typically provide limited analysis of whether these savings affect group-level reliability under variable acquisition conditions \cite{hooker2020what}. EcoFair is motivated by this gap and evaluates whether adaptive routing can jointly support deployment efficiency, degraded-input reliability, and group-level malignant-case performance in a vertically partitioned inference setting.

\section{Methodology}
\label{sec:methodology}

    EcoFair operationalises the central premise of this study: dependable medical inference in ad hoc and IoT environments must not be treated solely as a communication problem. Instead, it is a joint systems problem involving data-local representation exchange, adaptive computation, clinically informed decision support, and group-level evaluation under edge deployment constraints.
    
    \subsection{Partitioned Edge Architecture}

        EcoFair assumes an edge-inference setting in which image and tabular clients are logically separated and communicate learned representations to a fusion server. The current implementation simulates this topology within a centralised cross-validation pipeline in order to isolate the effect of routing, energy-aware escalation, and subgroup-sensitive evaluation. The architecture reduces direct raw-data exchange by keeping image pixels and tabular attributes local to their respective clients. Formal privacy mechanisms such as secure multiparty computation, homomorphic encryption, differential privacy, and secure aggregation are outside the current implementation. The threat scope is therefore centred on degraded or ambiguous sensor data, while representation-inversion, membership-inference, attribute-inference, Byzantine, and adversarial-embedding attacks remain future evaluation targets.
    
\begin{figure*}
            \centering
            \includegraphics[width=0.9\textwidth]{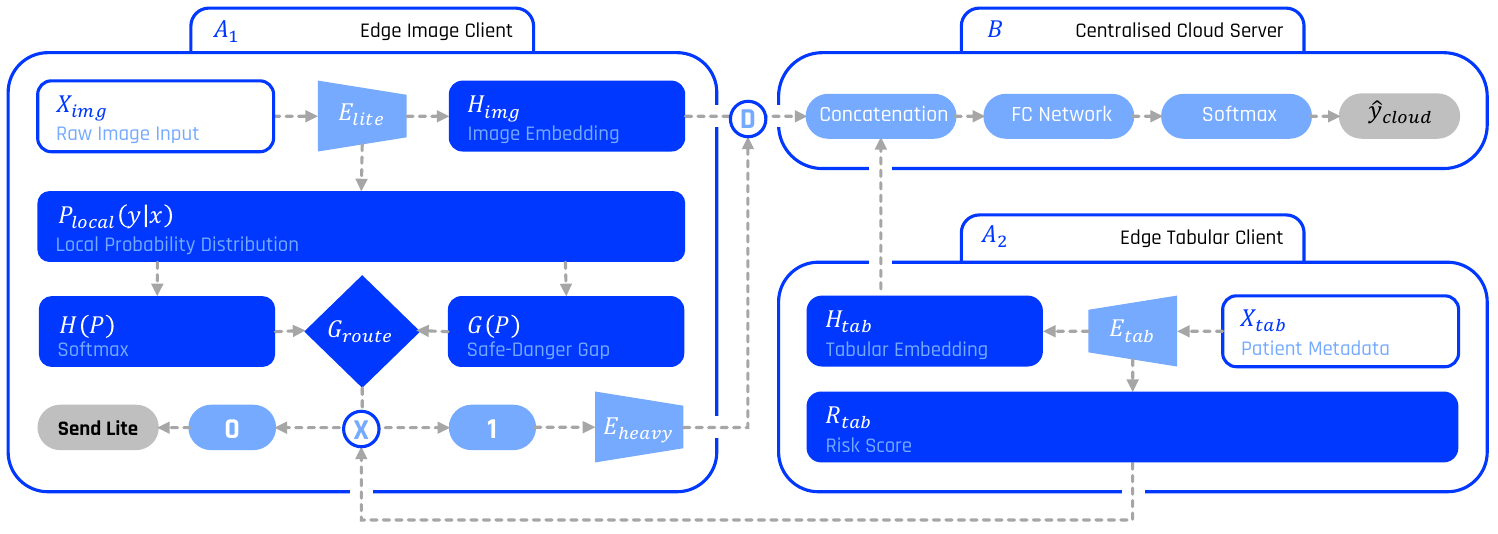}
            \caption{Architecture of the EcoFair partitioned edge inference framework. The image client (Party A1) computes routing signals from a lightweight encoder. To mitigate data degradation, a heavier encoder is adaptively activated only when necessary. Concurrently, the tabular client (Party A2) extracts embeddings and a metadata-derived risk score. Raw data remains strictly local; only modality-specific embeddings are transmitted to the server (Party B) for multimodal fusion.}
            \label{fig:architecture}
        \end{figure*}

        EcoFair is formulated as a simulated vertically partitioned medical inference architecture designed to preserve data locality while enabling reliable multimodal inference. The system follows a vertically partitioned topology with two local edge clients and one central server. Party A1 acts as the image client and processes raw dermatological images captured at the edge, where sensor data is frequently subject to noise and environmental degradation. Party A2 acts as the tabular client and processes patient metadata available locally, such as age and lesion localisation. Party B denotes the central server, which receives modality-specific embeddings and performs multimodal fusion to produce the final diagnostic prediction. The objective is to support collaborative inference under potentially adverse conditions without transmitting raw image pixels or raw tabular attributes beyond their originating edge nodes. Fig.~\ref{fig:architecture} illustrates the full EcoFair workflow.

        For empirical evaluation, EcoFair is studied in a simulated deployment setting using a centralised 5-fold cross-validation pipeline. This setup is intended to model adaptive routing behaviour under degraded-input conditions, data-local representation exchange, and edge resource economy, rather than to implement end-to-end federated training or measure real network latency. On the image side, Party A1 first computes a lightweight image embedding and the associated routing signals derived from the local predictive distribution. To handle potential data degradation safely, these signals determine whether the heavier image encoder is kept inactive or adaptively activated on the same client to compensate for local uncertainty. On the tabular side, Party A2 independently computes a tabular embedding together with a metadata-driven risk score. The server then receives the resulting modality-specific embeddings and performs multimodal fusion for the final prediction.

        The privacy-relevant design choice in EcoFair follows directly from this architectural constraint. Raw image and tabular inputs remain local to their respective clients, while only learned intermediate representations are transmitted for downstream fusion. This design reduces direct exposure of sensitive biometric and demographic information across potentially vulnerable or intermittent network links without relying on explicit cryptographic protocols or injected differential noise at inference time \cite{singh2019detailed}.

    \subsection{Degradation-Aware Inference Routing}
    
        A central design objective in EcoFair is to avoid unnecessary activation of the heavier image pathway while preserving clinically relevant diagnostic reliability. In a vertically partitioned deployment setting, the edge image client must first compute a lightweight representation for every input, but not every sample warrants the additional computational burden of a heavier encoder. EcoFair therefore introduces an adaptive inference routing mechanism that operates directly on the local predictive output of the lightweight image pathway. It decides whether the heavier image encoder should remain inactive or be dynamically activated for the same sample. This design targets adaptive resource management at the point of inference rather than during training, and is consistent with adaptive computation strategies proposed for resource-constrained intelligent systems \cite{han2022dynamic,li2020edge}.
        
        Let $x$ denote an input image processed by the lightweight image encoder $E_{\mathrm{lite}}$. The lightweight pathway produces an image embedding

        \begin{equation}
            h_{\mathrm{lite}}(x) = E_{\mathrm{lite}}(x),
        \end{equation}
        
        and a corresponding predictive distribution over $C$ classes
        
        \begin{equation}
            p(x) = \big[p_1(x), p_2(x), \dots, p_C(x)\big], \qquad \sum_{c=1}^{C} p_c(x)=1.
        \end{equation}
        
        The routing function is defined as a binary decision
        
        \begin{equation}
            G(x) \in \{0,1\},
        \end{equation}
        
        where $G(x)=0$ indicates that the lightweight image encoder alone is used, and $G(x)=1$ indicates that the heavier image encoder is additionally activated on the image client before embeddings are transmitted for multimodal fusion. Under both outcomes, raw image data remain local and only learned representations are sent to the server.

        The routing decision is derived from two lightweight statistics computed from the predictive distribution of the lightweight pathway. The first is the Shannon entropy
        
        \begin{equation}
            H(x) = -\sum_{c=1}^{C} p_c(x)\log p_c(x),
        \end{equation}
        
        which measures the spread of predictive mass across classes and is commonly used as a proxy for predictive uncertainty \cite{guo2017calibration,teerapittayanon2016branchynet}. In practical edge deployments, such uncertainty frequently stems from degraded, noisy, or ambiguous sensor inputs. Larger values of $H(x)$ indicate greater uncertainty and hence reduced confidence in the sufficiency of the lightweight pathway to handle the local data quality.

        The second statistic is a risk-oriented safe-to-danger probability gap. Let $\mathcal{S}$ denote the set of clinically lower-risk classes and let $\mathcal{D}$ denote the set of clinically higher-risk classes. The cumulative safe and danger probabilities are defined as
        
        \begin{equation}
            P_{\mathcal{S}}(x)=\sum_{s\in\mathcal{S}} p_s(x), \qquad
            P_{\mathcal{D}}(x)=\sum_{d\in\mathcal{D}} p_d(x),
        \end{equation}
        
        and the corresponding probability gap is
        
        \begin{equation}
            \Delta(x)=P_{\mathcal{S}}(x)-P_{\mathcal{D}}(x).
        \end{equation}
        
        A large positive value of $\Delta(x)$ indicates strong support for a lower-risk prediction, whereas values near zero indicate ambiguity between clinically safer and clinically riskier outcomes. To make this quantity directly usable as a routing signal, we define a gap-based ambiguity score so that larger values correspond to greater clinical ambiguity.
        
        \begin{equation}
            A_{\Delta}(x)=1-\big|\Delta(x)\big|,
        \end{equation}
        
        The final routing score combines global predictive uncertainty and safe to danger ambiguity,
        
        \begin{equation}
            R(x)=\lambda_H \,\widetilde{H}(x)+\lambda_{\Delta}\,A_{\Delta}(x),
            \label{eq:routing_score}
        \end{equation}
        
        where $\widetilde{H}(x)$ denotes a normalised entropy term and $\lambda_H,\lambda_{\Delta}\geq 0$ are weighting coefficients controlling the relative contribution of the two signals. In practice, EcoFair uses this combined score together with threshold-based gating,
        
        \begin{equation}
            G(x)=
            \begin{cases}
            1, & \text{if } R(x)>\tau_r,\\
            0, & \text{otherwise,}
            \end{cases}
        \end{equation}
        
        where $\tau_r$ is a calibrated routing threshold. Equivalently, this decision can be expressed as a logical trigger on the constituent signals,
        
        \begin{equation}
            G(x)=\mathbb{I}\!\left(\widetilde{H}(x)>\tau_H \;\; \lor \;\; A_{\Delta}(x)>\tau_{\Delta}\right),
        \end{equation}
        
        where $\tau_H$ and $\tau_{\Delta}$ are threshold parameters and $\mathbb{I}(\cdot)$ is the indicator function. This formulation makes the routing policy explicit: samples with high predictive uncertainty or a narrow safe-to-danger separation are treated as insufficiently reliable for lightweight-only processing. The routing rule is designed as a lightweight deployment heuristic. The additive score in Eq.~\ref{eq:routing_score} combines uncertainty and safe--danger ambiguity using quantities available after a single lightweight forward pass. This additive form prioritises auditability and low computational cost, while learned routing weights or nonlinear routing policies provide natural extensions for future work.
        
        The routing thresholds are calibrated within the cross-validation protocol using training-fold information only. For each fold, candidate entropy, safe--danger-gap, and metadata-risk thresholds are selected on the validation partition and then applied unchanged to the held-out fold, avoiding test-label leakage. Because threshold choice directly affects the energy--performance trade-off, Section~\ref{sec:routing_diagnostics} reports a post-hoc sensitivity sweep around the selected thresholds.

        In practical deployments where a fully labelled validation set is unavailable, analogous operating points could be selected using a small labelled pilot set, conservative target escalation rates, periodic audit as new labels become available, and site-specific recalibration under distribution shift.
        
        If $G(x)=0$, the image client transmits the lightweight image embedding for downstream fusion. If $G(x)=1$, the heavier image encoder $E_{\mathrm{heavy}}$ is additionally evaluated on the same client to produce

        \begin{equation}
            h_{\mathrm{heavy}}(x)=E_{\mathrm{heavy}}(x),
        \end{equation}
        
        In the implementation evaluated here, the server receives a single selected image representation rather than both image embeddings. The selected image representation is defined by the routing outcome as
        
        \begin{equation}
            h_{\mathrm{img}}(x)=
            \begin{cases}
            h_{\mathrm{lite}}(x), & G(x)=0,\\
            h_{\mathrm{heavy}}(x), & G(x)=1.
            \end{cases}
            \label{eq:selected_image_embedding}
        \end{equation}
        
        Because lightweight and heavyweight encoders may produce embeddings with different dimensionalities, the selected image embedding is mapped to a common fusion dimension using a pathway-specific projection layer,
        
        \begin{equation}
            z_{\mathrm{img}}(x)=\Pi_{m(x)}\!\left(h_{\mathrm{img}}(x)\right),
            \qquad
            m(x)=
            \begin{cases}
            \mathrm{lite}, & G(x)=0,\\
            \mathrm{heavy}, & G(x)=1.
            \end{cases}
            \label{eq:image_projection}
        \end{equation}
        
        The tabular embedding is similarly projected as
        
        \begin{equation}
            z_{\mathrm{tab}}(x)=\Pi_{\mathrm{tab}}\!\left(h_{\mathrm{tab}}(x)\right).
            \label{eq:tabular_projection}
        \end{equation}
        
        The server-side fusion operator then concatenates the projected image and tabular representations and applies the fusion classifier,
        
        \begin{equation}
            \hat{y}(x)=f_{\mathrm{B}}\!\left([z_{\mathrm{img}}(x);z_{\mathrm{tab}}(x)]\right),
            \label{eq:server_fusion}
        \end{equation}
        
        Here, \(\Pi_{m(x)}\) denotes the active image-pathway projection,
        \(\Pi_{\mathrm{tab}}\) denotes the tabular projection, 
        \([\cdot;\cdot]\) denotes concatenation, and 
        \(f_{\mathrm{B}}\) is the server-side fusion head. This formulation makes the routing-to-fusion transition explicit: non-escalated samples are fused using the projected lightweight image embedding, whereas escalated samples are fused using the projected heavyweight image embedding. By basing routing on quantities available from a single lightweight forward pass, EcoFair avoids the added computational burden of repeated stochastic inference procedures such as Monte Carlo Dropout, thereby making the routing step itself suitable for constrained edge deployment \cite{gal2016dropout,wang2020convergence}. Samples that are straightforward under the lightweight pathway avoid unnecessary heavy-model activation, while ambiguous, degraded, or clinically sensitive samples are selectively escalated. The resulting behaviour enables a direct trade-off between deployment economy, inference reliability, and group-level malignant-case performance, which is examined empirically in section~\ref{sec:experimental_evaluation}.

    \subsection{Rule-Based Metadata Risk Score}

        Predictive uncertainty alone is not always sufficient for reliable routing in edge environments, since neural models may still produce confident but incorrect predictions when processing noisy or degraded sensor data. To reduce this risk, EcoFair augments the routing mechanism with a rule-based metadata risk score derived from explicit patient metadata \cite{gaur2024building,hasani2022trustworthy}. This component provides a simple and auditable rule-based signal that operates alongside the image-based uncertainty measures.
        
        The tabular client, Party A2, processes metadata variables that are clinically associated with malignancy risk, specifically patient age and anatomical site. Lesion localisation is included not only as an anatomical descriptor, but also because body-site risk patterns in cutaneous melanoma are known to vary with differential ultraviolet exposure and related epidemiological factors, although this relationship is not uniform across all melanoma subtypes \cite{laskar2021risk,raimondi2020melanoma}. Rather than relying on latent image features, this component uses a deterministic rule-based formulation grounded in metadata-derived priors. Let \(a(x)\) denote the normalised age score for sample \(x\), and let \(\ell(x)\) denote the empirical localisation risk associated with the lesion site. The tabular risk score is defined as
        
        \begin{equation}
            R_{\mathrm{tab}}(x) = a(x)\,\ell(x),
        \end{equation}
        
        where both terms are scaled to the unit interval. In this form, larger values of \(R_{\mathrm{tab}}(x)\) correspond to samples with jointly elevated metadata-based risk.

        A more explicit formulation can be written as
        
        \begin{equation}
            a(x)=\frac{\mathrm{Age}(x)-A_{\min}}{A_{\max}-A_{\min}},
        \end{equation}
        
        and
        
        \begin{equation}
            \ell(x)=\frac{\mathrm{MalRate}(\mathrm{loc}(x))}{\max_{l}\mathrm{MalRate}(l)},
        \end{equation}
        
        where \(\mathrm{loc}(x)\) denotes the anatomical site of sample \(x\), and \(\mathrm{MalRate}(\cdot)\) is the empirical malignancy frequency estimated from the training data for each localisation category. This makes the metadata score directly interpretable as a metadata-driven prior on malignancy risk.
        
        The tabular score acts as an override on the image-side routing logic. If the lightweight edge model appears confident but the metadata-derived risk remains high, EcoFair treats this as a potential failure of the local image sensor to capture critical diagnostic features, activating the heavier encoder as a precaution. Formally, the risk-based override is defined as
        
        \begin{equation}
            G(x)=1 \qquad \text{if } R_{\mathrm{tab}}(x)\geq \tau_{\mathrm{risk}},
        \end{equation}
        
        where \(\tau_{\mathrm{risk}}\) is a calibrated risk threshold. Combined with the uncertainty-based routing conditions defined earlier, the final routing decision may therefore be written as
        
        \begin{equation}
            G(x)=\mathbb{I}\!\left(\widetilde{H}(x)>\tau_H \;\lor\; A_{\Delta}(x)>\tau_{\Delta} \;\lor\; R_{\mathrm{tab}}(x)\geq \tau_{\mathrm{risk}}\right).
        \end{equation}
        
        This override provides a metadata-based risk prior whose contribution is examined through risk-only and no-risk routing ablations in the experimental evaluation. In this sense, the metadata-risk component improves the interpretability of the routing process and supports selective escalation, while its empirical benefit is assessed rather than assumed \cite{tjoa2020survey,gaur2024building}.

    \subsection{Group-Level Routing Metrics}

        The final methodological component of EcoFair defines how group-level malignant-case behaviour is measured under adaptive routing. In standard training settings, optimisation is typically driven by aggregate objective functions, which can mask systematic disparities across demographic groups and clinically atypical cases \cite{oakden2020hidden}. In medical diagnosis, this may appear as reduced sensitivity for underrepresented cohorts even when overall performance remains acceptable. However, EcoFair departs from conventional fairness-inducing approaches as it does not impose fairness through an additional loss term or constrained optimisation objective during training. Instead, it examines whether subgroup-sensitive malignant-case performance changes post hoc through adaptive routing behaviour at inference time. The central idea is that samples associated with higher uncertainty, narrower safe to danger separation, or elevated metadata-derived risk are more likely to correspond to cases for which lightweight-only inference is vulnerable to data degradation. By selectively activating the heavier image pathway for such cases, EcoFair provides an operational mechanism for evaluating whether routing can support clinically important and subgroup-sensitive samples.
        
        Formally, let \(\mathcal{G}\) denote a set of demographic subgroups and let \(\mathrm{TPR}_g\) denote the true positive rate for subgroup \(g\in\mathcal{G}\) on malignant cases. EcoFair is evaluated using subgroup-sensitive criteria that examine both the mean malignant-case sensitivity across groups and the worst-group performance, serving as indicators of system robustness.
        
        \begin{equation}
            \mathrm{TPR}_{\mathrm{mean}}=\frac{1}{|\mathcal{G}|}\sum_{g\in\mathcal{G}}\mathrm{TPR}_g,
        \end{equation}
        
        and
        
        \begin{equation}            
            \mathrm{TPR}_{\mathrm{worst}}=\min_{g\in\mathcal{G}}\mathrm{TPR}_g.
        \end{equation}
        
        To quantify disparity, we further define a subgroup fairness gap as
        
        \begin{equation}
            \mathrm{Gap}_{\mathrm{TPR}}=\max_{g\in\mathcal{G}}\mathrm{TPR}_g-\min_{g\in\mathcal{G}}\mathrm{TPR}_g.
        \end{equation}
        
        Under this formulation, group-level reliability is evaluated empirically rather than enforced directly through parameter optimisation. The working hypothesis is that routing may improve subgroup dependability in selected configurations when escalation identifies samples for which the heavier encoder provides a useful corrective signal; conversely, subgroup disparity may worsen when escalation is unevenly distributed across groups or when the heavier encoder does not improve performance for the affected subgroup \cite{chen2021ethical}.

        Algorithm~\ref{alg:routing_logic} summarises the operational flow of EcoFair at inference time, showing how lightweight prediction, metadata-driven risk scoring, selective heavy-model activation, and server-side multimodal fusion are combined within a single resilient routing pipeline.
        
        \begin{algorithm}
            \caption{EcoFair Degradation-Aware Edge Routing and Server-Side Fusion}
            \label{alg:routing_logic}
            \begin{algorithmic}[1]
            \REQUIRE $x_{\mathrm{img}}$, $x_{\mathrm{tab}}$, $\tau_{\mathrm{risk}}, \tau_H, \tau_{\Delta}$
            \STATE $R_{\mathrm{tab}}(x) \leftarrow \text{ComputeRisk}(x_{\mathrm{tab}})$
            \STATE $p(x),\, h_{\mathrm{lite}}(x) \leftarrow \text{LiteEncoder}(x_{\mathrm{img}})$
            \STATE $H(x) \leftarrow -\sum_{c=1}^{C} p_c(x)\log p_c(x)$
            \STATE $\Delta(x) \leftarrow \sum_{s\in\mathcal{S}} p_s(x) - \sum_{d\in\mathcal{D}} p_d(x)$
            \STATE $A_{\Delta}(x) \leftarrow 1-\left|\Delta(x)\right|$
            \STATE $h_{\mathrm{tab}}(x) \leftarrow \text{TabularEncoder}(x_{\mathrm{tab}})$
            \IF{$R_{\mathrm{tab}}(x)\geq \tau_{\mathrm{risk}}$ \textbf{or} $H(x)\geq \tau_H$ \textbf{or} $A_{\Delta}(x)\geq \tau_{\Delta}$}
                \STATE $h_{\mathrm{img}}(x) \leftarrow \text{HeavyEncoder}(x_{\mathrm{img}})$
            \ELSE
                \STATE $h_{\mathrm{img}}(x) \leftarrow h_{\mathrm{lite}}(x)$
            \ENDIF
            \STATE Transmit $h_{\mathrm{img}}(x)$ and $h_{\mathrm{tab}}(x)$ to the server
            \STATE $z_{\mathrm{img}}(x) \leftarrow \Pi_{m(x)}\big(h_{\mathrm{img}}(x)\big)$ \STATE $z_{\mathrm{tab}}(x) \leftarrow \Pi_{\mathrm{tab}}\big(h_{\mathrm{tab}}(x)\big)$ \STATE $y_{\mathrm{final}} \leftarrow f_{\mathrm{B}}\big([z_{\mathrm{img}}(x);z_{\mathrm{tab}}(x)]\big)$
            \RETURN $y_{\mathrm{final}}$
            \end{algorithmic}
        \end{algorithm}

\section{Experimental Evaluation}
\label{sec:experimental_evaluation}
        
    The empirical study is designed to determine whether EcoFair delivers a tangible deployment advantage in practice by improving edge resource economy, preserving clinically relevant diagnostic performance, and examining group-level behaviour under adaptive heavy-model activation.

    \subsection{Evaluation Setup and Protocol}
        
        The evaluation uses three publicly available dermatology benchmarks: HAM10000 \cite{tschandl2018ham10000}, PAD-UFES-20 \cite{pacheco2020pad}, and BCN20000 \cite{hernandez2024bcn20000}. These datasets were selected because they collectively capture clinically relevant heterogeneity in image acquisition conditions, varying degrees of sensor data degradation, demographic composition, and label imbalance. The HAM10000 dataset serves as a controlled baseline of images collected under standard clinical conditions. In contrast, BCN20000 is used as a natural acquisition-variability and distribution-shift proxy, rather than as a formally quantified corruption benchmark. Furthermore, PAD-UFES-20 provides clinical images captured directly from smartphones, making it a useful proxy for heterogeneous edge-image acquisition, although it does not exhaustively model all physical sensor disturbances. Across all three benchmarks, the data reflects the inherent class imbalance patterns commonly observed in real-world dermatological edge deployments \cite{pacheco2021attention}. This makes them suitable for evaluating not only baseline predictive performance, but also routing behaviour and group-level performance under varying data quality.
        
\begin{figure*}
            \centering
            \includegraphics[width=0.9\textwidth]{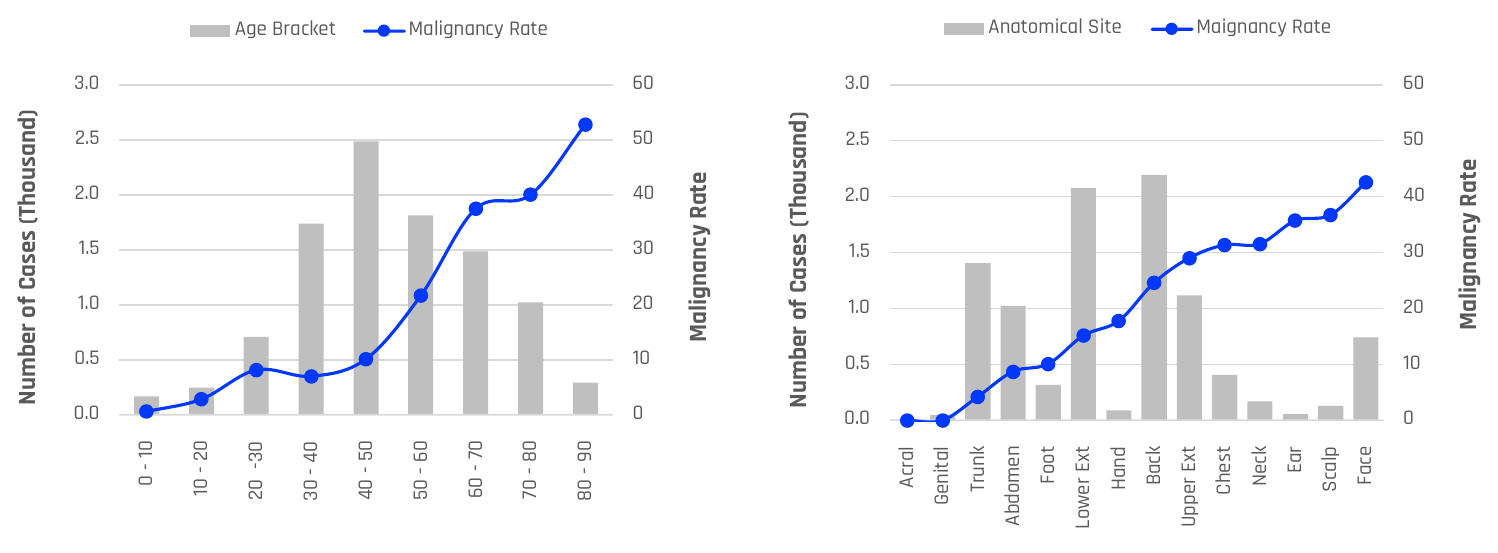}
            \caption{Empirical malignancy prevalence across patient age brackets (left) and anatomical localisations (right) in HAM10000. The non-uniform risk distribution across these metadata groups provides empirical support for the rule-based metadata risk score, which EcoFair uses to guide adaptive escalation when edge sensor data is degraded or ambiguous.}
            \label{fig:malignancy_distribution}
        \end{figure*}

        To support a principled analysis of deployment trade-offs, EcoFair does not rely on a single fixed image backbone. Instead, a broader screening stage is first used to profile multiple candidate encoders, after which three representative lite-heavy model pairs are selected for detailed framework-level evaluation. This design allows the experiments to examine EcoFair under moderate, strong, and extreme compute disparities between the lightweight baseline encoders and the heavyweight escalation pathways. The rule-based metadata risk score is calibrated from the training-fold metadata. Fig.~\ref{fig:malignancy_distribution} provides the empirical motivation for this metadata prior by showing that malignancy prevalence is not uniform across age brackets or anatomical sites within HAM10000. This supports the use of age and lesion localisation as metadata-driven routing signals for uncertain edge predictions.
        
    \subsection{Backbone Screening for Edge Deployments}
    
        To ensure that the architectural choices used in the EcoFair routing framework are principled rather than ad hoc, a comprehensive screening of ten widely used convolutional backbones was first conducted. 
        For each architecture, deployment-oriented characteristics including parameter count, model size, embedding dimensionality, inference latency, GPU memory footprint, and energy consumption per sample were empirically measured during feature extraction. 
        These measurements characterise the computational behaviour of each encoder independently of the routing mechanism and provide an objective basis for selecting representative model combinations for subsequent framework evaluation. To accurately quantify the computational cost of the feature extraction and inference phases, GPU energy consumption was dynamically tracked using pynvml, the official Python interface for the NVIDIA Management Library (NVML) \cite{pynvml}. This approach directly polls hardware-level power sensors, providing a highly accurate and widely validated standard for resource measurement in deep learning pipelines \cite{you2023zeus}. Energy measurements correspond to encoder-level image inference and feature extraction under the reported hardware configuration. They do not include radio transmission, server queuing, network retransmission, or full-device battery discharge. Therefore, the reported values should be interpreted as compute-side encoder energy estimates for comparing routing policies, not as end-to-end battery-life or full deployment-energy measurements.
        
        Table~\ref{tab:model_screening} summarises the screening statistics measured on HAM10000. Since encoder inference cost depends primarily on network architecture and input resolution rather than dataset content, the corresponding measurements on BCN20000 and PAD-UFES-20 were highly consistent and are therefore omitted for brevity. These additional results are available in the public implementation described in the Code Availability statement.

        \begin{table*}
            \centering
            \caption{Backbone screening results on HAM10000. Models are categorised into lightweight (top) and heavyweight (bottom) architectures, providing a structural division between baseline edge encoders and resource-intensive escalation pathways.}
            \label{tab:model_screening}
            \begin{tabular}{lcccccc}
            \toprule
            \multirow{2}{*}{\textbf{Model}} & \textbf{Parameters} & \textbf{Size} & \textbf{Embedding} & \textbf{Latency} & \textbf{Energy} & \textbf{GPU Memory} \\
            & \textbf{(M)}        & \textbf{(MB)} & \textbf{Dimension} & \textbf{(ms/Sample)} & \textbf{(J/Sample)} & \textbf{(MB)} \\
            \midrule
            MobileNetV3Small   & 0.94  & 3.58   & 576  & 143.50 & 0.179 & 465 \\
            MobileNetV2        & 2.26  & 8.61   & 1280 & 130.49 & 0.178 & 4499 \\
            MobileNetV3Large   & 2.99  & 11.43  & 960  & 167.67 & 0.209 & 1427 \\
            EfficientNetB0     & 4.05  & 15.45  & 1280 & 269.40 & 0.332 & 2451 \\
            NASNetMobile       & 4.27  & 16.29  & 1056 & 628.82 & 0.683 & 4499 \\
            \midrule
            ResNet50           & 23.59 & 89.98  & 2048 & 203.47 & 0.392 & 3457 \\
            DenseNet201        & 18.32 & 69.89  & 1920 & 682.11 & 0.844 & 15901 \\
            ResNet152V2        & 58.33 & 222.52 & 2048 & 566.06 & 0.933 & 9603 \\
            InceptionResNetV2  & 54.34 & 207.28 & 1536 & 682.81 & 1.238 & 15903 \\
            EfficientNetB6     & 40.96 & 156.25 & 2304 & 743.97 & 9.617 & 15903 \\
            \bottomrule
            \end{tabular}
        \end{table*}

        Based on the screening results in Table~\ref{tab:model_screening}, three representative model pairs were selected for the subsequent framework-level evaluation: Pair I (MobileNetV2 and ResNet50), Pair II (MobileNetV3Small and DenseNet201), and Pair III (MobileNetV3Small and EfficientNetB6). The objective of this selection is not to identify a universally optimal backbone, but to analyse the behaviour of the EcoFair routing mechanism under progressively stronger computational disparities between the routing pathways, using a small and interpretable set of representative configurations.

        These three pairs span complementary operating regimes. Pair I represents a practical deployment-oriented configuration with moderate computational contrast between lightweight and heavyweight image inference. Pair II introduces a substantially stronger heavy encoder to assess whether adaptive routing can recover difficult clinical cases caused by data degradation while retaining the efficiency advantages of a compact local model. Pair III provides an extreme compute contrast to examine the upper bound of resource economy achievable through selective escalation and to stress-test the robustness of the routing design under highly asymmetric deployment conditions. Taken together, these configurations allow the subsequent analysis to evaluate whether EcoFair can support diagnostic utility and group-level reliability while reducing total inference overhead across increasingly demanding edge scenarios.

    \subsection{Optimisation of Edge Resource Overhead}
    
        A primary deployment constraint in edge-based medical inference is the resource overhead associated with image processing under limited on-device capacity. In EcoFair, this cost is determined by whether inference proceeds with the lightweight image encoder alone or whether the heavier image encoder is adaptively activated for the same sample to compensate for local uncertainty. The routing mechanism, therefore, acts directly on the dominant source of variable computation in the edge client and provides a controllable trade-off between resource economy and inference reliability.
        
        \begin{table*}
            \centering
            \caption{Per-sample energy consumption and routing statistics across datasets. The evaluated configurations are Pair I (MobileNetV2 $\rightarrow$ ResNet50), Pair II (MobileNetV3Small $\rightarrow$ DenseNet201), and Pair III (MobileNetV3Small $\rightarrow$ EfficientNetB6). Reported values correspond to mean results across cross-validation folds.}
            \label{tab:energy_savings}
            \begin{tabular}{llccccc}
            \toprule
            \textbf{Dataset} & \textbf{Configuration} & \textbf{Lite} & \textbf{Heavy} & \textbf{EcoFair} & \textbf{Routing} & \textbf{Savings} \\
             &  & \textbf{(J/sample)} & \textbf{(J/sample)} & \textbf{(J/sample)} & \textbf{(\%)} & \textbf{vs. Heavy (\%)} \\
            \midrule
            \multirow{3}{*}{HAM10000}
            & Pair I   & 0.18 & 0.39 & 0.28 & 26.63 $\pm$ 2.82 & 28.09 $\pm$ 2.82 \\
            & Pair II  & 0.18 & 0.84 & 0.46 & 33.32 $\pm$ 1.96 & 45.43 $\pm$ 1.96 \\
            & Pair III & 0.18 & 9.62 & 3.09 & 30.04 $\pm$ 5.02 & 68.09 $\pm$ 5.02 \\
            \midrule
            \multirow{3}{*}{BCN20000}
            & Pair I   & 0.19 & 0.37 & 0.41 & 57.36 $\pm$ 7.75 & $-$10.09 $\pm$ 7.75 \\
            & Pair II  & 0.18 & 0.82 & 0.62 & 54.91 $\pm$ 4.12 & 24.27 $\pm$ 4.12 \\
            & Pair III & 0.18 & 9.37 & 6.01 & 57.10 $\pm$ 4.59 & 35.80 $\pm$ 4.59 \\
            \midrule
            \multirow{3}{*}{PAD-UFES-20}
            & Pair I   & 0.20 & 0.39 & 0.42 & 54.28 $\pm$ 2.66 & $-$7.73 $\pm$ 2.66 \\
            & Pair II  & 0.18 & 0.84 & 0.64 & 55.24 $\pm$ 3.31 & 23.86 $\pm$ 3.31 \\
            & Pair III & 0.18 & 9.61 & 5.58 & 56.04 $\pm$ 2.96 & 41.90 $\pm$ 2.96 \\
            \bottomrule
            \end{tabular}
        \end{table*}

        Table~\ref{tab:energy_savings} summarises the empirical per-sample energy consumption and routing statistics for the three selected model pairs across the evaluated datasets. The results show that the resource economy of EcoFair is strongly pair-dependent. On HAM10000, all three pairs achieve a reduction in energy consumption relative to their corresponding heavy baselines, with savings ranging from 28.09\% for MobileNetV2 and ResNet50 to 68.09\% for MobileNetV3Small and EfficientNetB6. This confirms that selective heavy-model activation can substantially reduce inference overhead when a meaningful computational gap exists between the lightweight baseline and the escalation pathway.
        
        The behaviour on BCN20000 and PAD-UFES-20 is more nuanced. For the DenseNet201 and EfficientNetB6 pairs, EcoFair continues to provide clear energy savings, although the gains are smaller than those observed on HAM10000. This occurs because the routing policy is triggered more frequently under stronger acquisition variability and distribution shift. By contrast, the MobileNetV2 and ResNet50 pair does not yield energy savings on BCN20000 or PAD-UFES-20. This indicates that when the computational contrast between the two encoders is relatively modest, frequent heavy-model activation can offset the efficiency benefit of the routing policy. This result highlights an important deployment trade-off: EcoFair may allocate additional computation under more difficult data regimes rather than maximising energy savings in every configuration.

        Taken together, these results show that EcoFair is most beneficial when the routing policy operates over a sufficiently large compute gap while retaining enough flexibility to escalate difficult or degraded cases. Its deployment benefit therefore depends on the interaction between routing behaviour, data quality, and backbone disparity.

        Beyond encoder-side energy, the same routing behaviour also affects communication and latency overhead. EcoFair transmits learned embeddings rather than raw images or metadata; based on the embedding dimensions in Table~\ref{tab:model_screening}, individual image embeddings range from approximately 2.25~KB to 9.00~KB per sample using float32 representations. Using the observed routing rates, the expected selected-embedding transmission is approximately 4.00--6.72~KB per sample across the evaluated configurations with complete artefacts, while a conservative setting that transmits both Lite and Heavy embeddings for escalated samples can reach approximately 9.57~KB per sample. These estimates do not include radio-layer overhead, packet retransmission, or protocol-specific metadata. In terms of execution time, every sample first incurs Lite-encoder inference and lightweight routing logic, while escalated samples additionally require Heavy-encoder computation. Table~\ref{tab:model_screening} reports encoder-level latency; however, the present setup does not measure end-to-end deployment latency variance under realistic wireless, server-load, or device-power conditions. The evaluation should therefore be interpreted as compute-side energy, encoder-latency, and representation-size analysis rather than a complete physical edge-network energy or latency study.

        \begin{figure*}
            \centering
            \includegraphics[width=0.9\textwidth]{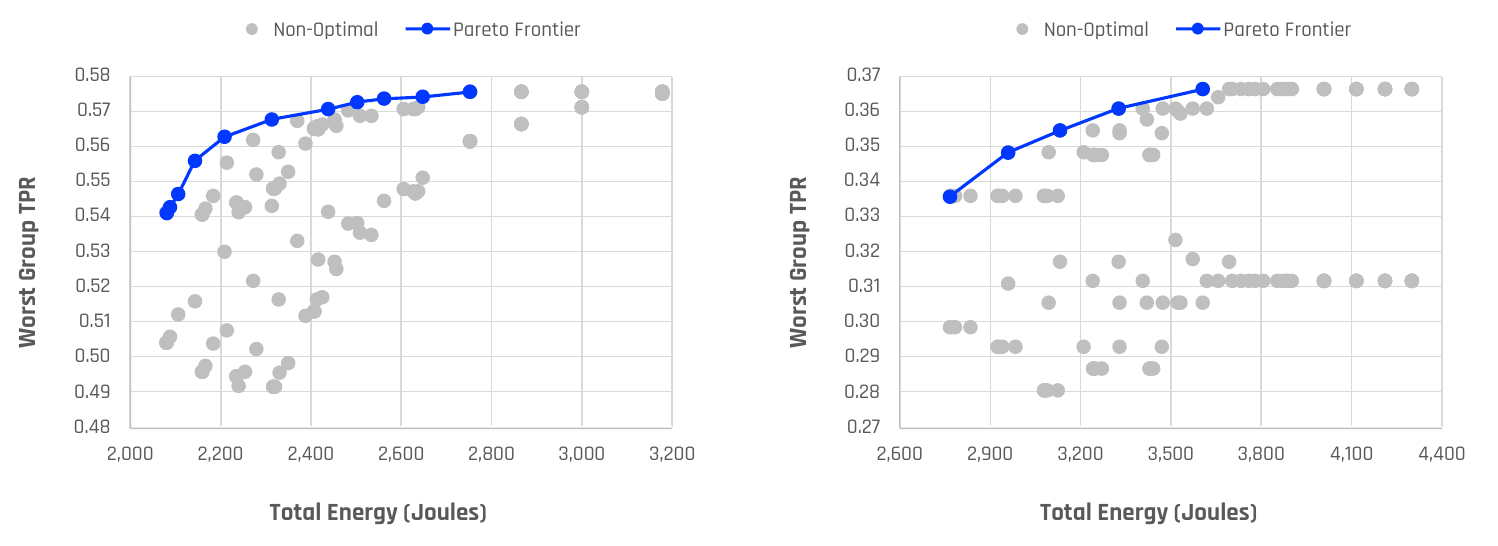}
            \caption{Pareto frontiers illustrating the trade-off between total edge energy consumption and worst-group diagnostic reliability for Pair I (MobileNetV2 and ResNet50). Left: On HAM10000, the framework shows a proportional trade-off, achieving high efficiency (down to 2,080 J) when local uncertainty is low. Right: On BCN20000, the operating regime shifts to a higher energy range (2,766--4,300 J), reflecting more frequent escalation under increased acquisition variability and distribution shift.}
            \label{fig:pareto_frontier}
        \end{figure*}

        Fig.~\ref{fig:pareto_frontier} illustrates this trade-off for the MobileNetV2 and ResNet50 pair on HAM10000, where the Pareto frontier identifies non-dominated operating points balancing total edge inference overhead against worst-group TPR on malignant classes. Additional Pareto frontiers for the remaining selected model pairs are available in the public implementation described in the Code Availability statement.

        The Pareto analysis serves as a deployment diagnostic that visualises feasible energy--dependability operating points. Each point represents a feasible operating setting with a different energy--dependability trade-off, and the preferred operating point would depend on the deployment constraint: battery-limited devices may favour lower-energy points, while higher-risk settings may justify greater escalation. Therefore, Fig.~\ref{fig:pareto_frontier} should be interpreted as an operational visualisation of the trade-off space rather than as a claim of globally optimal routing.

    \subsection{Classification Performance}

        Resource economy is only useful if it does not compromise deployment-oriented diagnostic performance. For this reason, the empirical evaluation considers not only overall class balance through Macro F1 and Balanced Accuracy, but also malignant-case recall, which is an important metric for malignant-case sensitivity. Table~\ref{tab:classification_performance} summarises these results across the three datasets and the selected model pairs.

        \begin{table*}
            \centering
            \caption{Classification performance across datasets. For each evaluated backbone pairing, results are reported for the standalone lightweight edge model (Lite), the standalone heavyweight escalation model (Heavy), and the adaptive routing framework (EcoFair). Reported values correspond to mean $\pm$ standard deviation across cross-validation folds. Bold values indicate the best result within each architecture block.}
            \label{tab:classification_performance}
            \begin{tabular}{llccc}
            \toprule
            \textbf{Dataset} & \textbf{Architecture} & \textbf{Macro F1} & \textbf{Balanced Accuracy} & \textbf{Malignant Recall} \\
            \midrule
            \multirow{9}{*}{HAM10000}
            & MobileNetV2 (Lite) & 0.5305 $\pm$ 0.0212 & 0.5932 $\pm$ 0.0324 & 0.5442 $\pm$ 0.0467 \\
            & ResNet50 (Heavy) & \textbf{0.5855 $\pm$ 0.0177} & \textbf{0.6431 $\pm$ 0.0227} & \textbf{0.5998 $\pm$ 0.0308} \\
            & EcoFair & 0.5629 $\pm$ 0.0124 & 0.6224 $\pm$ 0.0230 & 0.5519 $\pm$ 0.0299 \\
            \cmidrule(lr){2-5}
            & MobileNetV3Small (Lite) & 0.5521 $\pm$ 0.0165 & 0.6244 $\pm$ 0.0165 & \textbf{0.6168 $\pm$ 0.0342} \\
            & DenseNet201 (Heavy) & \textbf{0.5837 $\pm$ 0.0118} & 0.6338 $\pm$ 0.0288 & 0.5802 $\pm$ 0.0206 \\
            & EcoFair & 0.5774 $\pm$ 0.0222 & \textbf{0.6440 $\pm$ 0.0262} & 0.5882 $\pm$ 0.0477 \\
            \cmidrule(lr){2-5}
            & MobileNetV3Small (Lite) & 0.5568 $\pm$ 0.0205 & 0.6285 $\pm$ 0.0235 & \textbf{0.5793 $\pm$ 0.0335} \\
            & EfficientNetB6 (Heavy) & 0.5468 $\pm$ 0.0182 & 0.6141 $\pm$ 0.0171 & 0.5714 $\pm$ 0.0142 \\
            & EcoFair & \textbf{0.5784 $\pm$ 0.0163} & \textbf{0.6506 $\pm$ 0.0157} & 0.5850 $\pm$ 0.0251 \\
            \midrule
            \multirow{9}{*}{BCN20000}
            & MobileNetV2 (Lite) & 0.3750 $\pm$ 0.0355 & 0.3988 $\pm$ 0.0399 & 0.4152 $\pm$ 0.0538 \\
            & ResNet50 (Heavy) & 0.3816 $\pm$ 0.0261 & 0.4094 $\pm$ 0.0284 & \textbf{0.4587 $\pm$ 0.0281} \\
            & EcoFair & \textbf{0.3908 $\pm$ 0.0321} & \textbf{0.4165 $\pm$ 0.0362} & 0.4430 $\pm$ 0.0427 \\
            \cmidrule(lr){2-5}
            & MobileNetV3Small (Lite) & 0.3819 $\pm$ 0.0157 & 0.4437 $\pm$ 0.0237 & 0.4706 $\pm$ 0.0293 \\
            & DenseNet201 (Heavy) & 0.3748 $\pm$ 0.0177 & 0.4012 $\pm$ 0.0186 & 0.4632 $\pm$ 0.0402 \\
            & EcoFair & \textbf{0.3940 $\pm$ 0.0200} & \textbf{0.4566 $\pm$ 0.0261} & \textbf{0.4750 $\pm$ 0.0387} \\
            \cmidrule(lr){2-5}
            & MobileNetV3Small (Lite) & 0.3922 $\pm$ 0.0297 & 0.4311 $\pm$ 0.0235 & \textbf{0.4732 $\pm$ 0.0380} \\
            & EfficientNetB6 (Heavy) & 0.3774 $\pm$ 0.0193 & 0.4123 $\pm$ 0.0207 & 0.4594 $\pm$ 0.0234 \\
            & EcoFair & \textbf{0.4015 $\pm$ 0.0323} & \textbf{0.4419 $\pm$ 0.0218} & 0.4658 $\pm$ 0.0345 \\
            \midrule
            \multirow{9}{*}{PAD-UFES-20}
            & MobileNetV2 (Lite) & 0.5986 $\pm$ 0.0170 & 0.6090 $\pm$ 0.0064 & 0.5261 $\pm$ 0.0370 \\
            & ResNet50 (Heavy) & \textbf{0.6228 $\pm$ 0.0331} & \textbf{0.6360 $\pm$ 0.0279} & 0.5365 $\pm$ 0.0479 \\
            & EcoFair & 0.6157 $\pm$ 0.0253 & 0.6243 $\pm$ 0.0286 & \textbf{0.5384 $\pm$ 0.0643} \\
            \cmidrule(lr){2-5}
            & MobileNetV3Small (Lite) & 0.6128 $\pm$ 0.0177 & 0.6647 $\pm$ 0.0250 & 0.6267 $\pm$ 0.0676 \\
            & DenseNet201 (Heavy) & 0.6147 $\pm$ 0.0265 & 0.6315 $\pm$ 0.0144 & 0.5447 $\pm$ 0.0425 \\
            & EcoFair & \textbf{0.6351 $\pm$ 0.0231} & \textbf{0.6722 $\pm$ 0.0274} & \textbf{0.6283 $\pm$ 0.0452} \\
            \cmidrule(lr){2-5}
            & MobileNetV3Small (Lite) & 0.6147 $\pm$ 0.0415 & 0.6549 $\pm$ 0.0402 & 0.6280 $\pm$ 0.0638 \\
            & EfficientNetB6 (Heavy) & 0.6016 $\pm$ 0.0285 & 0.6292 $\pm$ 0.0202 & 0.5561 $\pm$ 0.0261 \\
            & EcoFair & \textbf{0.6374 $\pm$ 0.0357} & \textbf{0.6756 $\pm$ 0.0310} & \textbf{0.6450 $\pm$ 0.0435} \\
            \bottomrule
            \end{tabular}
        \end{table*}
        
        The results show that EcoFair behaves as an adaptive deployment mechanism whose benefit depends on the interaction between dataset difficulty and backbone disparity. On HAM10000, EcoFair consistently remains competitive and achieves the strongest overall classification profile for the two larger-compute-gap pairs. Pair~I represents a relatively modest compute disparity, whereas Pairs~II and III introduce substantially stronger heavyweight pathways. These larger gaps provide the routing mechanism with a greater opportunity to safely alter the deployment trade-off. In particular, the MobileNetV3Small and DenseNet201 configuration, alongside the EfficientNetB6 pairing, yield the best Balanced Accuracy values, indicating that adaptive escalation can improve class-balanced performance without requiring the heavyweight pathway for every sample.

        On BCN20000 and PAD-UFES-20, the degradation-aware behaviour of the framework becomes evident. For the DenseNet201 and EfficientNetB6 pairs, EcoFair either matches or exceeds both baselines across most reported metrics. This is most prominent on PAD-UFES-20, where EcoFair achieves the strongest performance under both larger-gap pairings and improves malignant recall relative to the corresponding heavy baselines. Because these datasets contain stronger acquisition variability and distribution shift, this performance supports the value of adaptive routing under degraded or variable edge data. By contrast, the MobileNetV2 and ResNet50 pair produces more modest gains, which is consistent with the smaller representational separation between those two encoders. 
        
        Taken together, these results support the intended role of EcoFair as an adaptive deployment mechanism rather than a uniformly dominant classifier. Its strongest performance arises when routing operates over a sufficiently meaningful gap between baseline and escalation pathways, allowing difficult or degraded cases to be reliably recovered while preserving the economy of local edge processing.

    \subsection{Group-Level Evaluation}

        Group-level evaluation is important in edge deployment because acceptable average performance can still conceal systematic failure on clinically relevant subgroups. In this study, subgroup behaviour is evaluated primarily through the Equal Opportunity perspective, which examines parity in True Positive Rate (TPR) across demographic subgroups for clinically important malignant cases \cite{hardt2016equality}. Under this criterion, subgroup behaviour is more favourable when worst-group sensitivity improves and the TPR gap across subgroups decreases, rather than when only average performance increases on the dominant population.
        
\begin{figure*}
            \centering
            \includegraphics[width=0.9\textwidth]{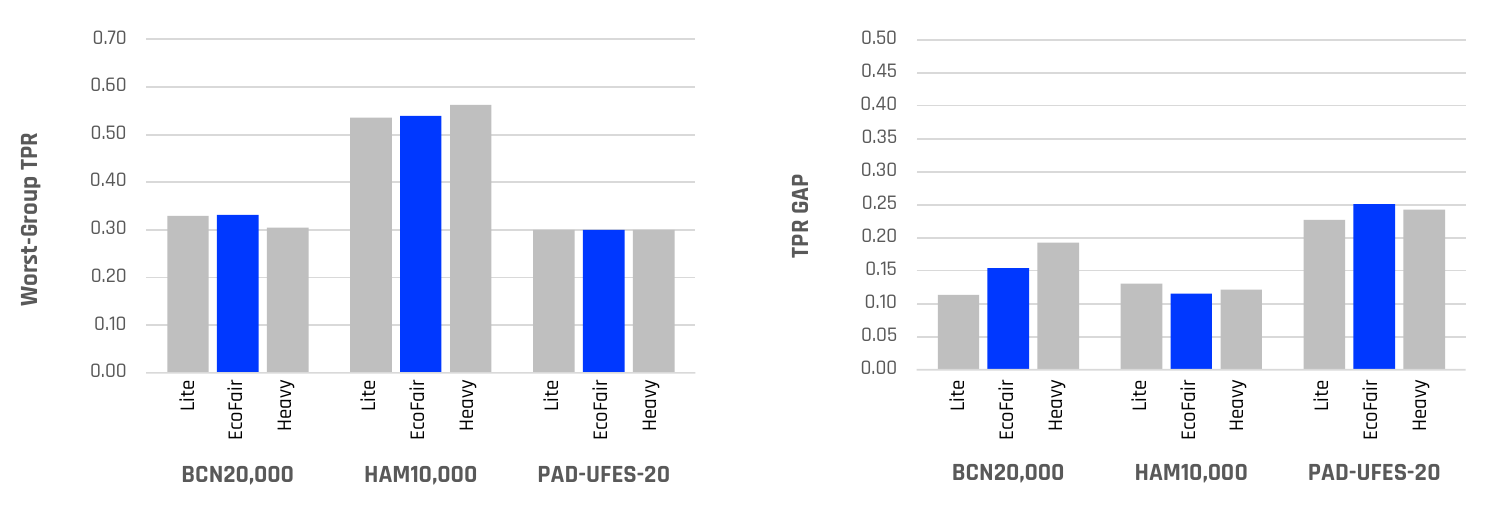}
            \caption{Group-level evaluation for Pair I (MobileNetV2 and ResNet50) across the three benchmark datasets. The left plot shows worst-group TPR on malignant classes, and the right plot shows the corresponding TPR gap across demographic subgroups.}
            \label{fig:fairness_rescue}
        \end{figure*}

        EcoFair does not enforce fairness through a modified training objective or an explicit fairness penalty. Instead, it investigates whether subgroup dependability changes post hoc through adaptive routing behaviour at inference time. Samples associated with higher uncertainty, narrower safe to danger separation, or elevated metadata-derived risk are more likely to be difficult for lightweight-only inference and are therefore more likely to trigger selective heavy-model activation. As shown in Fig.~\ref{fig:fairness_rescue} and summarised in Table~\ref{tab:fairness_delta}, the subgroup behaviour of EcoFair is configuration-dependent. The analysis evaluates whether selectively escalating samples that are more likely to be ambiguous or degraded can improve group-level malignant-case performance relative to the baselines.

        \begin{table}
            \centering
            \caption{Subgroup dependability changes relative to Lite and Heavy baselines. The evaluated pairs are Pair I (MobileNetV2 and ResNet50), Pair II (MobileNetV3Small and DenseNet201), and Pair III (MobileNetV3Small and EfficientNetB6). Positive values indicate improvement in the corresponding metric, while negative values indicate deterioration. Bold values indicate the strongest improvement within each dataset block.}
            \label{tab:fairness_delta}
            \setlength{\tabcolsep}{4pt} 
            \begin{tabular}{llrrrr}
            \toprule
            \multirow{2}{*}{\textbf{Data}} & \multirow{2}{*}{\textbf{Pair}} & \multicolumn{2}{c}{\textbf{vs. Heavy}} & \multicolumn{2}{c}{\textbf{vs. Lite}} \\
            \cmidrule(lr){3-4} \cmidrule(lr){5-6}
            & & \textbf{$\Delta$ WG TPR} & \textbf{$\Delta$ Gap} & \textbf{$\Delta$ WG TPR} & \textbf{$\Delta$ Gap} \\
            \midrule
            \multirow{3}{*}{HAM} 
            & I   & $-$0.023 & 0.006 & \textbf{0.004} & \textbf{0.015} \\
            & II  & \textbf{0.010} & \textbf{0.129} & $-$0.046 & $-$0.029 \\
            & III & 0.008 & $-$0.087 & $-$0.017 & $-$0.011 \\
            \midrule
            \multirow{3}{*}{BCN} 
            & I   & 0.027 & \textbf{0.039} & 0.002 & $-$0.040 \\
            & II  & 0.031 & 0.013 & \textbf{0.015} & 0.000 \\
            & III & \textbf{0.042} & 0.023 & $-$0.022 & $-$0.024 \\
            \midrule
            \multirow{3}{*}{PAD} 
            & I   & 0.000 & $-$0.009 & 0.000 & $-$0.024 \\
            & II  & $-$0.200 & $-$0.275 & $-$0.100 & $-$0.103 \\
            & III & \textbf{0.200} & \textbf{0.112} & 0.000 & $-$0.023 \\
            \bottomrule
            \end{tabular}
        \end{table}
        
        The results indicate that this behaviour is configuration-dependent and mixed. For Pair I, EcoFair provides the most stable subgroup reliability profile across datasets, yielding modest improvements in worst-group TPR relative to the lightweight baseline on HAM10000 and BCN20000, and reducing the disparity gap relative to the heavyweight baseline on BCN20000. This makes Pair I the clearest example of a practically balanced routing policy, while still showing that subgroup effects must be interpreted per dataset and configuration.

        The remaining pairs help clarify both the limits and failure cases of the framework. Pair II shows that routing can produce dependability gains in some settings, particularly when the heavier model provides a useful corrective signal, but it also worsens subgroup dependability on PAD-UFES-20 relative to both baselines. Pair III demonstrates that a larger compute gap can still support strong recovery behaviour against the heavyweight baseline in selected cases, although this comes with less consistent behaviour relative to the lightweight baseline. Taken together, these results show that group-level effects are configuration-dependent. Selective heavy-model activation can improve worst-group reliability in selected settings, but it can also worsen disparity when escalation is unevenly distributed across demographic groups or when the heavier encoder does not correct the errors affecting the worst-performing subgroup \cite{wiens2019roadmap}.
        
        Having evaluated energy, classification, and group-level outcomes, the final analysis examines why the routing policy behaves this way by isolating the contribution of each routing signal and testing sensitivity to threshold choice.

    \subsection{Routing Diagnostics and Threshold Sensitivity}
    \label{sec:routing_diagnostics}
    
        To isolate the contribution of each routing signal, we performed a post-hoc diagnostic analysis using the stored out-of-fold predictions from all datasets and backbone pairings. We compared Lite-only inference, Heavy-only inference, entropy-only routing, safe--danger-gap-only routing, metadata-risk-only routing, entropy-plus-gap routing without metadata risk, and the full EcoFair policy. The entropy-only and safe--danger-gap-only variants also act as simple confidence- and ambiguity-based adaptive cascade baselines. These baselines are directly comparable within the current post-hoc routing setting because they operate on the same stored Lite and Heavy predictions. By contrast, early-exit architectures such as BranchyNet \cite{teerapittayanon2016branchynet} train intermediate classifiers inside a single neural network, whereas EcoFair routes between independently trained lightweight and heavyweight encoders in a vertically partitioned multimodal setting. This distinction motivates treating early-exit networks, adaptive computation time, mixture-of-experts routing, and learned conditional-computation policies as future retrained dynamic-inference comparators. No model was retrained; alternative routing decisions were reconstructed from the saved routing components.

        Table~\ref{tab:routing_ablation_compact} shows that no single signal is universally dominant. Entropy-only routing is conservative, while safe--danger-gap and entropy-plus-gap routing provide stronger balanced-accuracy gains. Metadata-risk-only routing improves malignant recall relative to Lite-only inference, but with smaller balanced-accuracy gains. Full EcoFair achieves the strongest average malignant recall among the adaptive variants and matches the best balanced accuracy, at the cost of a higher routing rate. These results support interpreting EcoFair as a practical routing heuristic rather than an optimal policy.
        
        \begin{table*}
            \centering
            \caption{Routing signal ablation averaged across the nine dataset--backbone configurations. Energy is reported as expected image-encoder energy per sample and averaged over configurations with available energy artefacts.}
            \label{tab:routing_ablation_compact}
            \scriptsize
            \setlength{\tabcolsep}{4pt}
            \renewcommand{\arraystretch}{1.05}
            \resizebox{\textwidth}{!}{%
            \begin{tabular}{lrrrrrrr}
            \toprule
            \textbf{Routing variant} &
            \textbf{Route (\%)} &
            \textbf{Energy (J)} &
            \textbf{Bal. Acc.} &
            \textbf{Macro F1} &
            \textbf{Mal. Recall} &
            \textbf{WG TPR} &
            \textbf{TPR Gap} \\
            \midrule
            Lite only & 0.00 & 0.181 & 0.556 & 0.512 & 0.536 & 0.404 & 0.181 \\
            Heavy only & 100.00 & 2.721 & 0.555 & 0.523 & 0.528 & 0.389 & 0.192 \\
            Entropy-only routing & 3.86 & 0.204 & 0.559 & 0.515 & 0.538 & 0.410 & 0.177 \\
            Safe--danger-gap-only routing & 42.74 & 1.317 & 0.571 & 0.529 & 0.534 & 0.392 & 0.194 \\
            Metadata-risk-only routing & 9.91 & 0.307 & 0.558 & 0.515 & 0.542 & 0.404 & 0.186 \\
            Entropy + gap without metadata risk & 44.32 & 1.324 & 0.573 & 0.531 & 0.537 & 0.398 & 0.191 \\
            Full EcoFair routing & 47.63 & 1.364 & 0.573 & 0.532 & 0.542 & 0.398 & 0.195 \\
            \bottomrule
            \end{tabular}%
            }
        \end{table*}
        
        We also assessed threshold sensitivity, uncertainty quality, and statistical significance. Threshold multipliers from 0.8 to 1.2 showed the expected energy--routing trade-off: routing rates decreased as thresholds increased, while balanced accuracy varied only modestly across HAM10000, BCN20000, and PAD-UFES-20. Lite-model entropy yielded AUROC values of 0.687--0.818 for Lite-model error detection, supporting its use as a useful but imperfect uncertainty signal. Paired fold-level tests further showed that some balanced-accuracy gains were statistically supported, particularly on HAM10000 and selected BCN20000 configurations, while several malignant-recall and PAD-UFES-20 differences were not significant. Accordingly, small numerical gains are treated as descriptive unless supported by the statistical tests.

\section{Discussion}
\label{sec:discussion}

    In sensor-degraded edge-AI settings, EcoFair contributes a runtime adaptation mechanism for the specific disturbance class of degraded or ambiguous sensor inputs. Rather than assuming that all samples should receive identical computation, the system treats compute as an adaptive reliability resource. This is relevant for medical IoT scenarios where lightweight inference is efficient for routine samples but may be insufficient for uncertain or shifted inputs.
    
    The empirical results position EcoFair as a degradation-aware routing framework rather than a universally dominant static classifier. Its systemic benefit depends heavily on the interaction between the adaptive routing policy, environmental data degradation, and the computational disparity between the lightweight and heavyweight edge pathways. The strongest results are observed when routing operates over a meaningful compute gap. Here, selective heavy-model activation can improve classification balance and group-level metrics in selected configurations while reducing inference overhead under standard operating conditions.
    
    A second critical finding is that subgroup dependability emerges as an empirical outcome of the routing policy rather than as a constraint imposed during training. EcoFair does not enforce subgroup parity through constrained optimisation or loss modification. Instead, adaptive escalation improves worst-group malignant-case behaviour in some configurations, but can also leave subgroup gaps unchanged or worsen them when the heavy pathway does not correct the relevant subgroup errors. This configuration-dependent behaviour is now explicitly reflected in the interpretation of Table~\ref{tab:fairness_delta}.
    
    The energy results further show that EcoFair's savings are configuration- and data-dependent. On more variable or shifted data, such as BCN20000, the routing rule escalates more frequently, which reduces net energy savings for some backbone pairs. This reflects the intended trade-off of the framework: additional computation is allocated when lightweight predictions appear less reliable.

    At the same time, the present results define a bounded validation scope. The evaluation focuses on encoder-side energy, adaptive routing, classification behaviour, and group-level outcomes under benchmark variation. Deployment factors such as packet loss, network partitions, radio-transmission energy, protocol overhead, latency variance under real network load, adversarial cyber manipulation, malicious clients, and physical battery-discharge tests remain outside the current evaluation. EcoFair is therefore best viewed as a routing and evaluation layer for degraded-input handling within a broader secure edge-AI pipeline.

\section{Future Work}
\label{sec:future_work}

    While these results demonstrate the degradation-aware behaviour of the routing logic, translating these gains to physical ad hoc networks introduces additional systemic variables that remain open for future investigation.
    
    First, the current evaluation is conducted in a simulated vertically partitioned inference setting rather than on a physically distributed topology. A natural next step is to deploy EcoFair on physical edge hardware and embedded medical IoT platforms to measure end-to-end network latency, latency variance, communication bandwidth bottlenecks, radio-transmission energy, and actual battery depletion under practical operating conditions.
    
    Second, the current framework relies on the availability and quality of local tabular metadata. In real-world edge environments, sensors may fail or transmit incomplete records. Future work should therefore examine the reliability of the metadata-based risk component under missing, corrupted, or biased metadata signals, investigating more reliable strategies for sensor validation and uncertainty-aware tabular reasoning.

    Third, the datasets used in this study provide representative variation in image quality, acquisition conditions, and distribution shift, but future work should extend this evaluation with controlled disturbance protocols. This includes measured blur, low illumination, compression, occlusion, packet loss, missing metadata, and explicit OOD scoring.

    Finally, the present results show that the effectiveness of EcoFair is heavily pair-dependent. Future work should extend the framework to include stronger adaptive-computation baselines, such as BranchyNet-style early-exit networks, adaptive computation time, mixture-of-experts routing, and learned conditional-computation policies. These comparisons require architectural modification, retraining, or additional optimisation objectives, and would allow EcoFair to be assessed against fully trained dynamic-inference alternatives as well as post-hoc routing policies.

\section{Conclusion}
\label{sec:conclusion}

    The viability of next-generation healthcare IoT depends on resolving the tension between data locality, limited edge resources, and diagnostic dependability. In this work, we introduced EcoFair, a vertically partitioned inference architecture designed to navigate these competing objectives through adaptive routing. The framework keeps raw image and tabular inputs local to their originating clients while transmitting learned representations for server-side fusion.
    
    By combining lightweight-first inference with metadata-derived risk scoring, EcoFair establishes a degradation-triggered escalation mechanism. Under settings with a sufficient computational gap between encoders, the architecture reduces unnecessary heavy-model activation and lowers encoder-side energy consumption. Under more variable or shifted data conditions, the routing mechanism escalates uncertain inputs more frequently, reflecting a deployment trade-off between edge resource economy and inference reliability in the evaluated benchmark setting. 
    
    The results show that group-level malignant-case behaviour can be evaluated, and in selected configurations improved, through deployment-time routing; however, these effects remain configuration-dependent rather than universally guaranteed. Overall, EcoFair provides an energy-aware and subgroup-sensitive routing framework for ad hoc medical edge inference, supporting the evaluation of trade-offs between resource economy, degraded-input reliability, and group-level malignant-case performance.

\section*{Code Availability}

    The implementation is publicly available at \url{https://github.com/mociatto/EcoFair}. The repository includes the experimental pipelines, feature-extraction workflows, routing modules, evaluation scripts, and additional-analysis workflow for ablation, threshold sensitivity, uncertainty quality, statistical testing, and energy/transmission summaries.

\section*{Acknowledgment}
    Funding: This work was supported by the UK National Edge AI Hub [Grant Agreement EP/Y028813/1].

\bibliographystyle{unsrt}
\bibliography{references.bib}

\end{document}